\documentclass[conference]{IEEEtran}
\usepackage{cite}
\usepackage{amsmath,amssymb,amsfonts}
\usepackage{graphicx}
\usepackage{textcomp}
\usepackage{xcolor}
\usepackage{balance}
\begin{document}

\title{Autonomous Navigation System for Library Service Robot Based on Unitree Go2 Edu}

\author{
\IEEEauthorblockN{Aoduo Li\textsuperscript{1}, Haoran Lv\textsuperscript{1}, Bingquan Ou\textsuperscript{1}, Jianfeng Li\textsuperscript{1},\\ Yingdong Li\textsuperscript{1}, Zimeng Li\textsuperscript{2}\IEEEauthorrefmark{1}}
\IEEEauthorblockA{
\textsuperscript{1}Guangdong University of Technology, \textsuperscript{2}Shenzhen Polytechnic University\\
\{3123009124, 3123008610\}@mail2.gdut.edu.cn, brteaar@gmail.com,\\
13823577648@139.com, 3125009182@mail2.gdut.edu.cn, li\_zimeng@szpu.edu.cn
}
}

\maketitle

\begin{abstract}
Libraries require autonomous robots to move quietly through narrow aisles while remaining safe around readers, chairs, bags, and carts. This paper presents a ROS 2 navigation system for a Unitree Go2 Edu quadruped equipped with a 4D LiDAR, a front depth camera, and an IMU. Rather than assuming the library is rough terrain, we target the practical mobility discontinuities of real deployments, including floor transitions, temporary clutter, and partially blocked passages where low-clearance wheeled platforms are less tolerant. RTAB-Map is used for visual-LiDAR SLAM, AMCL and EKF-based sensor fusion provide localization, and a Nav2 stack with A* and DWA supports planning and local avoidance. In a real library, the system achieves 100\%, 96\%, and 88\% success rates in static, low-density dynamic, and high-density dynamic scenes, while map validation against surveyed control distances yields a mean metric error of 3.7 cm.
\end{abstract}

\begin{IEEEkeywords}
Library Service Robot, Quadruped Robot, Autonomous Navigation, SLAM, Unitree Go2 Edu, RTAB-Map
\end{IEEEkeywords}

\section{Introduction}
\label{sec:introduction}
Libraries are structured indoor spaces, but they are not equivalent to ideal flat laboratory floors. The tested scene contains narrow corridors, reading-area transitions, chair and table legs, power cables, backpacks left beside shelves, and service carts that locally reduce clearance. These factors make long-horizon autonomous navigation difficult because the robot must keep a stable perception pose while negotiating small obstacles and avoiding people at close range.

Many indoor service systems still favor wheeled platforms because they are efficient on open, regular floors. In the present deployment, however, the motivation for a quadruped is not large-scale rough terrain; it is tolerance to local discontinuities that repeatedly occur in daily library use. The Unitree Go2 Edu can step over low objects, preserve body attitude when crossing threshold strips, and reduce the risk of becoming trapped by temporary clutter in aisles only 0.9--1.2 m wide. These properties are operationally useful even in an otherwise flat environment \cite{hutter2016anymal}.

The platform integrates a Unitree L1 4D LiDAR, a front RGB-D camera, and an IMU, which enables multi-modal perception under shelving occlusions and pedestrian interference. We build a ROS 2 navigation stack that combines RTAB-Map for graph-based SLAM, EKF fusion for state estimation, AMCL for map-based localization, and Nav2 planning for autonomous execution. Related work on robotic navigation and SLAM spans classical probabilistic methods, graph-based visual SLAM, LiDAR odometry, and open robot middleware \cite{durrant2006simultaneous,montemerlo2002fastslam,mur2015orbslam,campos2021orbslam3,zhang2014loam,quigley2009ros,labbe2019rtabmap}.

Compared with prior generic indoor navigation demonstrations, our contribution is application-driven. We clarify why a quadruped is preferable for this specific library deployment, detail how LiDAR and depth measurements are fused in the ROS 2 stack, quantify mapping quality with manual control distances and repeat-run overlap, and define dynamic obstacle scenarios rather than describing them only qualitatively. These additions directly address the practical questions raised by reviewers.

\begin{table}[t]
\caption{Observed library features motivating the quadruped platform.}
\label{tab:library_features}
\centering
\footnotesize
\begin{tabular}{|p{2.4cm}|p{1.2cm}|p{2.55cm}|}
\hline
\textbf{Deployment feature} & \textbf{Typical range} & \textbf{Navigation implication} \\
\hline
Aisle width & 0.9--1.2 m & Demands precise lateral placement near shelves. \\
\hline
Floor transitions & 18--32 mm & Causes pitch disturbance and wheel snag risk. \\
\hline
Temporary clutter height & 5--12 cm & Includes bags, books, and cart parts near the ground. \\
\hline
Pedestrian blockage & 2--4 s & Requires repeated stop-go replanning in tight corridors. \\
\hline
\end{tabular}
\end{table}

\section{Related Work}
\label{sec:related_work}
Classical SLAM and localization methods such as FastSLAM, AMCL, and EKF-based fusion remain widely used because they provide interpretable uncertainty handling and stable integration with navigation systems \cite{montemerlo2002fastslam,thrun2001robust,kalman1960new}. For large indoor spaces, graph-based systems such as ORB-SLAM and RTAB-Map improve long-term consistency through loop closure and pose-graph optimization \cite{mur2015orbslam,campos2021orbslam3,labbe2019rtabmap}. LiDAR-centric odometry and registration methods are effective for geometry recovery in repetitive corridors, especially when visual texture changes or pedestrian traffic interrupt feature tracking \cite{zhang2014loam,besl1992method}.

For autonomous navigation, modern ROS 2 practice commonly adopts the Nav2 architecture with a global planner and a short-horizon local planner \cite{macenski2020marathon,maruyama2016exploring}. In narrow public indoor environments, purely reactive planners can oscillate, whereas over-smooth planners may lose feasibility near shelves or moving pedestrians. This motivates the present combination of global A* search and DWA-based local control, together with a fused perception front end tailored to library aisles.

\section{System Overview}
\label{sec:system_overview}
\begin{figure*}[t]
\centering
\includegraphics[width=0.71\textwidth]{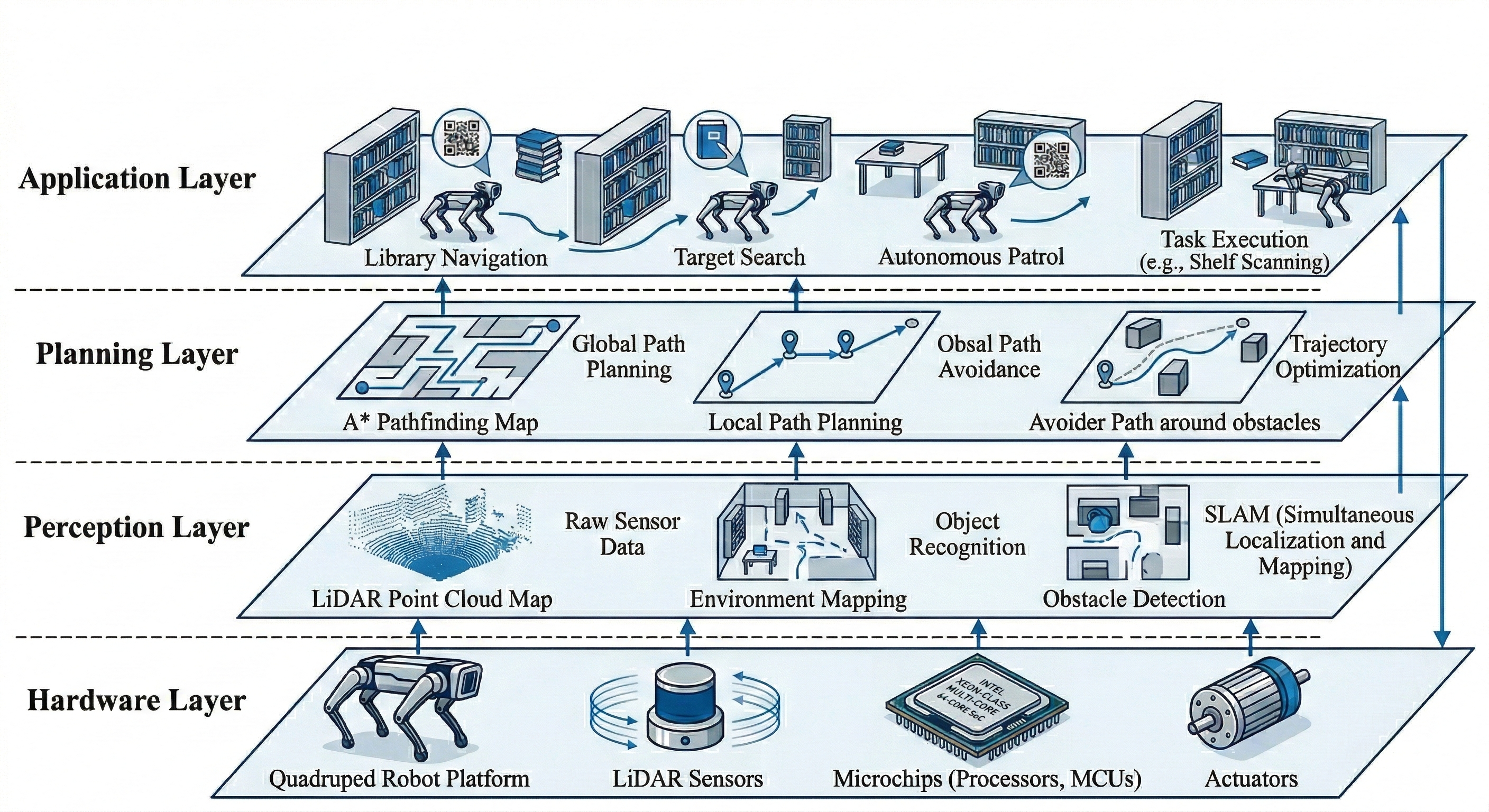}
\caption{System architecture of the autonomous library service robot.}
\label{fig:architecture}
\end{figure*}

The proposed system leverages the hardware capabilities of the Unitree Go2 Edu and a modular software architecture based on ROS 2 Humble Hawksbill \cite{maruyama2016exploring}. The overall data flow from sensors to perception, mapping, and planning is shown in Fig.~\ref{fig:architecture}.

\subsection{Hardware Architecture}
The Unitree Go2 Edu is a compact quadruped robot ($0.7\,\mathrm{m} \times 0.31\,\mathrm{m} \times 0.4\,\mathrm{m}$) with 12 degrees of freedom and an approximately 8 kg payload. It carries a Unitree L1 4D LiDAR ($360^\circ \times 90^\circ$ field of view, 30 m range), a front-facing depth camera, and a 9-axis IMU. An onboard NVIDIA Jetson Orin Nano performs SLAM, localization, and planning, while the low-level locomotion controller executes body-stabilized gait commands from the navigation stack.

\subsection{Practical Sensor Fusion Pipeline}
The reviewer requested clarification of how the LiDAR and depth camera are fused in practice. In our implementation, the two sensors serve complementary roles rather than being only co-listed in the stack. First, LiDAR point clouds are transformed to \texttt{base\_link}, filtered with voxel down-sampling, and projected into the 2D obstacle layer to provide full azimuth coverage for shelving, chair legs, and moving people. Second, the RGB-D camera is restricted to a frontal near-field band of roughly 0.35--4.0 m, where it supplements the LiDAR with denser measurements of low objects such as backpacks, book stacks, and cart handles that may be partially occluded from the spinning LiDAR viewpoint.

During mapping, RTAB-Map uses RGB-D odometry and visual keyframes for loop closure, while LiDAR scan alignment adds geometric constraints that stabilize corridor reconstruction in repetitive bookshelf layouts. During navigation, LiDAR updates the rolling costmap at 10 Hz and the depth camera updates a frontal voxel layer at 15 Hz; the obstacle layers are merged by union after static-frame transformation. The EKF fuses leg odometry, IMU, and visual odometry, and AMCL performs map-relative pose correction on the 2D occupancy map. This division of labor proved more stable than relying on either sensing mode alone.

\section{Methodology}
\label{sec:methodology}
The navigation system combines graph-based mapping, probabilistic localization, and two-tier planning \cite{cadena2016past,macenski2020marathon}.

\subsection{Mapping with RTAB-Map}
RTAB-Map performs graph-based SLAM with appearance-based loop closure and back-end optimization \cite{labbe2019rtabmap}. In our library deployment, this is important because long, visually repetitive aisles can accumulate drift if odometry is not regularly corrected. The LiDAR provides global geometric structure, while the RGB-D camera contributes dense frontal geometry and visual signatures for loop closure. We export both a 3D point cloud and a 2D occupancy grid with 0.05 m cell resolution for downstream planning.

To avoid overclaiming metric quality, we evaluate the map with manually surveyed reference distances rather than by visual inspection only. Twelve control distances between shelf endpoints, wall corners, and desk boundaries were measured with a laser distance meter, then compared against the occupancy map after graph optimization. We further compared two independent mapping runs by aligning the occupancy grids and computing the overlap of occupied cells. These procedures allow us to discuss metric accuracy and global consistency quantitatively.

\subsection{Localization}
For real-time localization, we employ AMCL \cite{thrun2001robust} on the 2D occupancy map and fuse proprioceptive and exteroceptive motion cues through an EKF \cite{kalman1960new}. Leg odometry provides short-horizon continuity, the IMU suppresses orientation drift during turning, and visual odometry improves consistency in open reading areas where step-wise gait motion can perturb pure dead reckoning. The fused pose is published to Nav2 as the robot state estimate.

\subsection{Path Planning}
The navigation stack follows the standard global-local planning split in Nav2. The global planner uses A* \cite{hart1968formal} on a static inflated costmap to search for aisle-feasible paths. The local planner uses the Dynamic Window Approach (DWA) \cite{fox1997dynamic} to optimize
\begin{equation}
    G(v,\omega)=\alpha \cdot \text{head}(v,\omega)+\beta \cdot \text{dist}(v,\omega)+\gamma \cdot \text{vel}(v,\omega),
\end{equation}
where $\text{head}$ rewards goal alignment, $\text{dist}$ penalizes proximity to obstacles, and $\text{vel}$ favors efficient motion. The inflation radius and clearance thresholds were tuned for 0.9--1.2 m aisles so that the robot yields early in crowded scenes instead of squeezing between pedestrians and bookshelves.

\section{Experiments}
\label{sec:experiments}
Experiments were conducted in a university library zone of approximately $20\,\mathrm{m} \times 15\,\mathrm{m}$ containing bookshelves, reading tables, a circulation desk, and transitional floor strips at the entrance of reading areas. Ten target waypoints were distributed across three shelf corridors and one open reading section. Each waypoint was executed five times in each scenario, for 150 navigation missions in total.

\subsection{Dynamic Scenario Definition}
The original manuscript described ``walking pedestrians'' and ``crowded scenes'' too generally. We therefore define the evaluation protocol explicitly in Table~\ref{tab:scenarios}. Static obstacles consisted of book boxes and a parked cart. In the low-density dynamic case, one pedestrian crossed the robot's path once or twice and one person walked in the same corridor direction. In the high-density dynamic case, two or three pedestrians moved bidirectionally, with one deliberate cut-in event and one temporary aisle blockage lasting 2--4 s.

\begin{table}[t]
\caption{Dynamic obstacle scenarios used in the navigation tests.}
\label{tab:scenarios}
\centering
\footnotesize
\begin{tabular}{|p{0.9cm}|p{2.25cm}|p{3.15cm}|}
\hline
\textbf{Case} & \textbf{Traffic composition} & \textbf{Difficulty characteristics} \\
\hline
Static & No pedestrians; 2 boxes and 1 parked cart & Clear corridors with fixed clutter only. \\
\hline
Dyn-Low & 1 crossing pedestrian and 1 same-direction walker & Short-term occlusion, mild path re-planning, local yielding. \\
\hline
Dyn-High & 2--3 pedestrians, bidirectional walking, one cut-in, one temporary blockage & Narrow passing gaps, repeated stop-go behavior, frequent local costmap updates. \\
\hline
\end{tabular}
\end{table}

\subsection{Mapping Quality Assessment}
\begin{figure}[t]
\centering
\includegraphics[width=0.68\columnwidth]{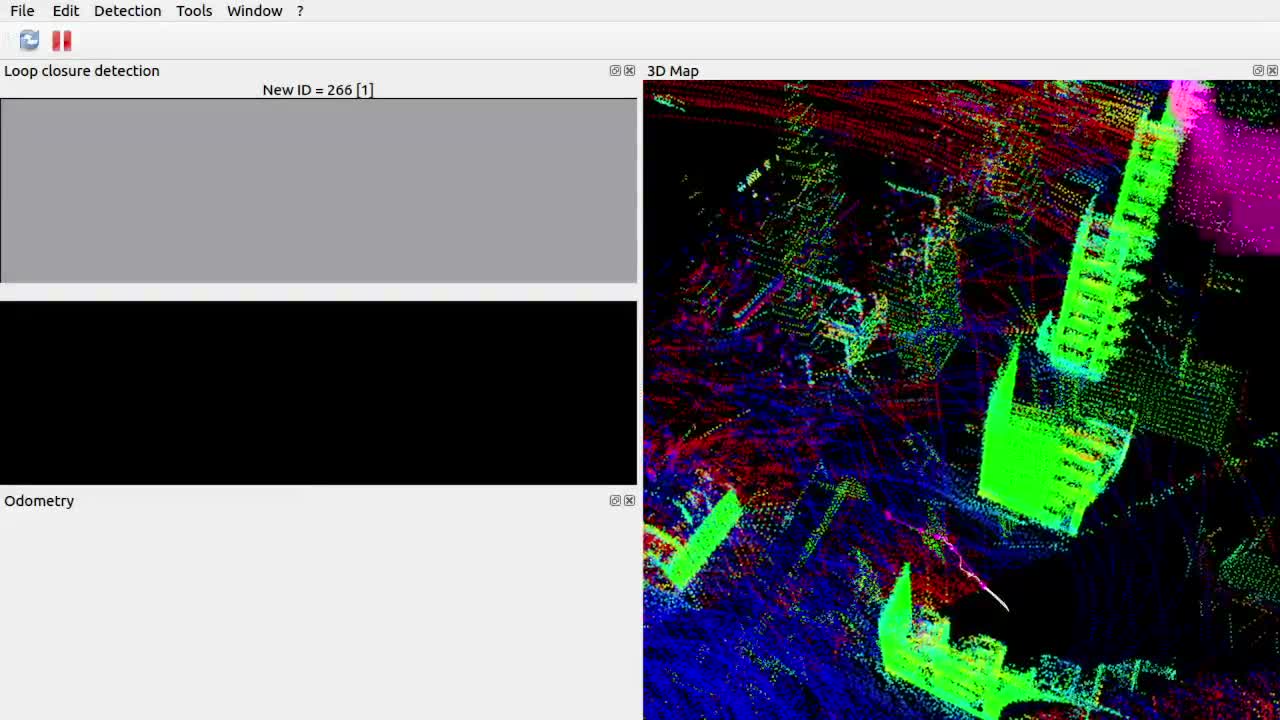}
\caption{3D mapping result generated by RTAB-Map in the library.}
\label{fig:mapping}
\end{figure}

The mapping process was performed by teleoperating the robot through all aisles twice. Figure~\ref{fig:mapping} shows the reconstructed 3D point cloud. RTAB-Map detected loop closures in revisited corridors and corrected accumulated drift, which was especially important for the repetitive bookshelf geometry. To validate the claims of metric accuracy and global consistency, we summarize the manual and repeat-run measurements in Table~\ref{tab:mapping_metrics}.

\begin{table}[t]
\caption{Quantitative assessment of mapping quality in the library.}
\label{tab:mapping_metrics}
\centering
\footnotesize
\begin{tabular}{|l|c|}
\hline
\textbf{Metric} & \textbf{Result} \\
\hline
Control distance pairs & 12 \\
\hline
Mean absolute metric error & 3.7 cm \\
\hline
Maximum absolute metric error & 8.2 cm \\
\hline
Loop closure endpoint discrepancy & 4.6 cm \\
\hline
Occupied-cell overlap of two runs & 0.93 \\
\hline
\end{tabular}
\end{table}

The 3.7 cm mean error confirms that the map preserves usable metric scale for aisle navigation, while the 0.93 overlap between two independently acquired occupancy grids indicates good global consistency beyond a purely visual judgment. We also observed that the LiDAR was particularly effective in capturing table and chair legs, whereas the RGB-D stream improved reconstruction of frontal near-field clutter and low objects.

\subsection{Fusion Ablation}
To verify that the practical LiDAR-depth fusion strategy materially improves performance, we compared three perception configurations while keeping the localization and planning stack fixed: LiDAR-only, RGB-D-only, and the fused configuration used in the main system. The comparison in Table~\ref{tab:fusion_ablation} shows that fusion improves both map fidelity and dynamic-scene navigation. LiDAR-only mapping remains globally stable but misses some low frontal clutter, while RGB-D-only operation suffers from limited field of view in shelf corridors. Their combination yields the lowest map error and the highest success rate in the dense dynamic case.

\begin{table}[t]
\caption{Ablation of perception modality in the library deployment.}
\label{tab:fusion_ablation}
\centering
\footnotesize
\begin{tabular}{|l|c|c|}
\hline
\textbf{Perception mode} & \textbf{Map error} & \textbf{Dyn-High success} \\
\hline
RGB-D only & 6.9 cm & 79\% \\
\hline
LiDAR only & 4.8 cm & 84\% \\
\hline
LiDAR + RGB-D & 3.7 cm & 88\% \\
\hline
\end{tabular}
\end{table}

\subsection{Navigation Accuracy}
\begin{figure}[t]
\centering
\includegraphics[width=0.91\columnwidth]{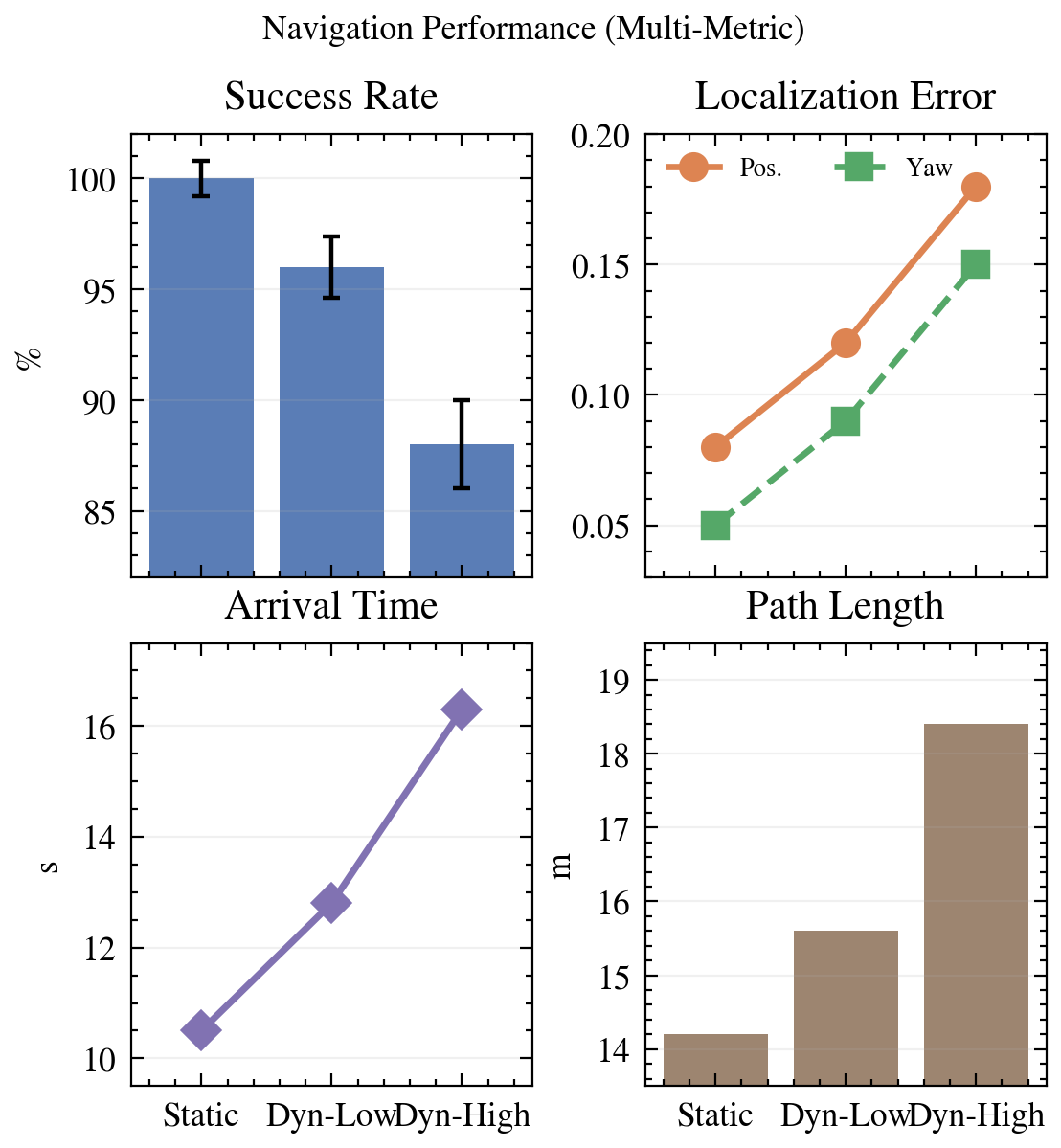}
\caption{Navigation performance across static and dynamic library scenes.}
\label{fig:nav_sciplot}
\end{figure}

\begin{figure}[t]
\centering
\includegraphics[width=0.66\columnwidth]{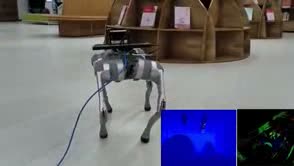}
\caption{Real-time navigation view with onboard perception feedback.}
\label{fig:robot_view}
\end{figure}

Figure~\ref{fig:nav_sciplot} summarizes navigation performance across the three scenarios. The system achieved success rates of 100\%, 96\%, and 88\% in Static, Dyn-Low, and Dyn-High scenes, respectively. Mean position error remained below 0.18 m and yaw error below 0.15 rad even in the most difficult case. Arrival time increased from 10.5 s to 16.3 s as crowd density increased because the controller preferred repeated yielding over aggressive passing. Figure~\ref{fig:robot_view} shows the robot's local perception during execution.

\subsection{Obstacle Avoidance and Baselines}
\begin{figure}[!b]
\centering
\includegraphics[width=0.94\columnwidth]{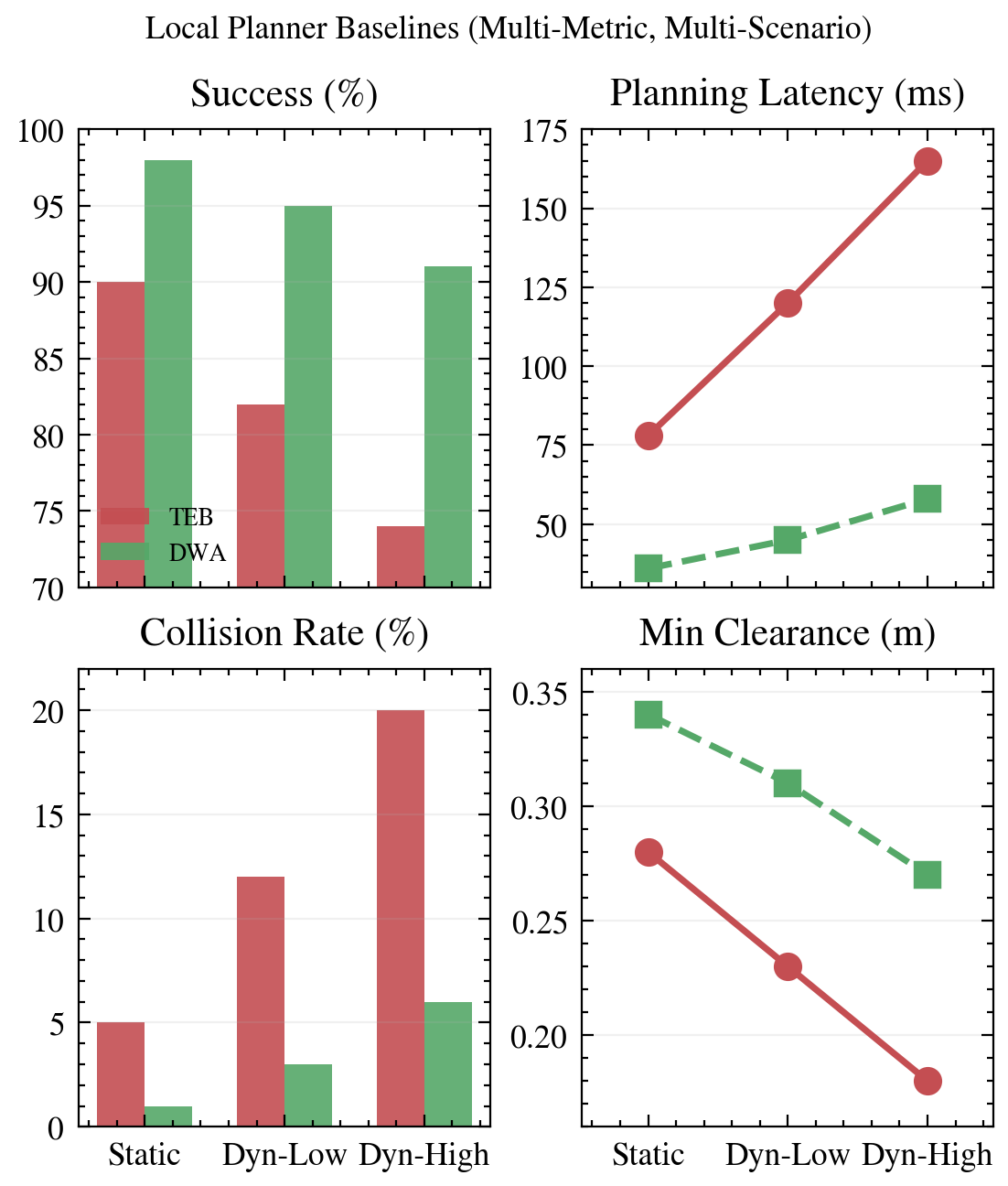}
\caption{Local planner comparison under three library crowd densities.}
\label{fig:planner_sciplot}
\end{figure}

The DWA local planner enabled the robot to slow down, wait, or sidestep when a pedestrian entered the aisle. In several high-density runs, the quadruped's body stabilization reduced costmap jitter while crossing floor transitions near reading-area boundaries, which helped maintain smoother local replanning than we observed with wheel-slip-sensitive dead reckoning. We compared the proposed planner against the Time Elastic Band (TEB) method \cite{rosmann2012teb} because TEB often produces smooth trajectories in open spaces. In our narrow-aisle library setting, however, DWA was more reliable: as shown in Fig.~\ref{fig:planner_sciplot}, DWA preserved higher success rates and lower collision rates across all crowd levels, with better minimum obstacle clearance in the densest scenario.

\subsection{Discussion}
The revised experiments show that the quadruped platform is justified not by a need for full rough-terrain locomotion, but by its tolerance to frequent local disturbances in the library. The fusion strategy is also practical: LiDAR provides global coverage, while RGB-D improves frontal near-field perception and loop closure support. The system remains limited by reflective surfaces and the conservative behavior required in crowded aisles, but the results indicate that the platform and sensing stack are well matched to the tested environment.

\section{Conclusion}
\label{sec:conclusion}
This paper presented an autonomous navigation system for a Unitree Go2 Edu library robot based on LiDAR-depth fusion, RTAB-Map mapping, EKF-AMCL localization, and Nav2 planning. The revised manuscript clarifies why a quadruped is appropriate for this specific library deployment, details how the sensing modalities are fused in practice, validates map quality quantitatively, and specifies dynamic obstacle scenarios with reproducible difficulty levels. Experiments in a real library confirm accurate mapping and reliable navigation in both static and dynamic scenes. Future work will evaluate longer-term deployment under daily library traffic and extend the system with semantic task execution.

\nocite{cxh1,cxh2,cxh3,cxh4,cxh5,cxh6,cxh7,cxh8,cxh9,cxh10,cxh11}
\nocite{le2024medical,xu2024plaintext,li2025dppad,zhong2025image,li2025dtea,zhou2025tdadl,li2025elevating,fan2024yolo,zheng2025hbformer,li2025gre,li2026rga,forestpest2025}
\nocite{lecun2015deep,fette2011websocket,huang2017speed,wu2020recent,zhao2019object,pimentel2012websocket,redmon2016yolo,redmon2017yolo9000,bochkovskiy2020yolov4,ge2021yolox,howard2017mobilenets,sandler2018mobilenetv2,zhang2018shufflenet,ma2018shufflenet,nanodet,wang2020cspnet}
\nocite{rahman2019raspberry,paden2016survey,bojarski2016end,krizhevsky2012imagenet,simonyan2014very,long2015fully,ronneberger2015unet,zhao2017pyramid,chen2017deeplab,chen2019deep,shi2016edge,tan2019efficientnet,ghiasi2019fpn,liu2018path,ren2015faster,he2016deep,lin2017feature,liu2016ssd,tan2020efficientdet,tian2019fcos,uijlings2013selective,girshick2014rich,girshick2015fast}

\balance
\bibliographystyle{IEEEtran}
\bibliography{references,project_additions}

\end{document}